\documentclass[11pt]{article}

\usepackage[]{acl}
\usepackage{times}
\usepackage{latexsym}
\usepackage{multirow}
\usepackage{tcolorbox}
\usepackage[T1]{fontenc}
\usepackage[utf8]{inputenc}
\usepackage{nccmath}
\usepackage{adjustbox}
\usepackage{microtype}
\usepackage{enumitem}
\usepackage{inconsolata}
\usepackage{amssymb}
\usepackage{amsmath} 
\usepackage{graphicx}
\usepackage{subcaption}
\usepackage{xcolor}
\usepackage[table]{xcolor}  % [table] 选项是关键！
\usepackage{colortbl}  
\usepackage{mathtools}
\usepackage{diagbox}
\usepackage{booktabs} 
\renewcommand{\thefootnote}{\fnsymbol{footnote}}

\title{Compressing then Matching: An Efficient Pre-training Paradigm for Multimodal Embedding}

\author{Da Li \textsuperscript{\rm 1,2}\footnotemark[1]\space\space 
Yuxiao Luo \textsuperscript{\rm 3}\footnotemark[1]\space\space
Keping Bi\textsuperscript{\rm 1,2 }\footnotemark[2]\space\space 
Jiafeng Guo\textsuperscript{\rm 1,2}\footnotemark[2]\space\space 
Wei Yuan\textsuperscript{\rm 3}\footnotemark[3]  \space\space \\
\textbf{Biao Yang}\textsuperscript{\rm 3}\space\space 
\textbf{Yan Wang}\textsuperscript{\rm 3}\space\space 
\textbf{Fan Yang}\textsuperscript{\rm 3}\space\space
\textbf{Tingting Gao}\textsuperscript{\rm 3}\space\space
\textbf{Guorui Zhou}\textsuperscript{\rm 3}\space\space\\
\textsuperscript{\rm 1}State Key Laboratory of AI Safety, Institute of Computing Technology, \\
Chinese Academy of Sciences \\
\textsuperscript{\rm 2}University of Chinese Academy of Sciences\,\,
\textsuperscript{\rm 3} Kuaishou Technology \\
\{lida21s, bikeping,guojiafeng\}@ict.ac.cn \\
\{luoyuxiao, yuanwei05, yangbiao, wangyan33, yangfan\}@kuaishou.com \\
 }

\newcommand{\methodname}{CoMa} 
\begin{document}
\maketitle
\begin{abstract}
Multimodal Large Language Models advance multimodal representation learning by acquiring transferable semantic embeddings, thereby substantially enhancing performance across a range of vision-language tasks, including cross-modal retrieval, clustering, and classification. An effective embedding is expected to comprehensively preserve the semantic content of the input while simultaneously emphasizing features that are discriminative for downstream tasks. Recent approaches demonstrate that MLLMs can be adapted into competitive embedding models via large-scale contrastive learning, enabling the simultaneous optimization of two complementary objectives. We argue that the two aforementioned objectives can be decoupled: a comprehensive understanding of the input enables the embedding model to achieve superior performance on downstream tasks via contrastive learning.
In this paper, we propose \textbf{\methodname}, a compressed pre-training phase, which serves as a warm-up stage for contrastive learning. 
Experiments demonstrate that with only a small amount of pre-training data, we can transform an MLLM into a competitive embedding model. \methodname~achieves new state-of-the-art results among MLLMs of comparable size on the MMEB, realizing optimization in both efficiency and effectiveness. Our project is available at https://github.com/Trustworthy-Information-Access/CoMa.
\end{abstract}
\footnotetext[1]{Contributed equally}
\footnotetext[2]{Corresponding authors}
\footnotetext[3]{Project leader}
% \footnotetext[1]{Contributed equally}
% \footnotetext[2]{Corresponding authors}
\renewcommand{\thefootnote}{\arabic{footnote}}
\setcounter{footnote}{0}
\section{Introduction}
Multimodal embedding is a core research area in artificial intelligence. It integrates heterogeneous data from modalities such as text, images, audio, and video to build cross-modal representations with rich semantics. These representations play an essential role in enabling key downstream tasks like image-text retrieval~\citep{Wang_2024,tang2025missingtargetrelevantinformationprediction}, Retrieval Augmented Generation (RAG)~\citep{yao2025expandrteachingdenseretrievers,gao2024retrievalaugmentedgenerationlargelanguage}, and Visual Question Answering (VQA)~\citep{garderes-etal-2020-conceptbert,chun2021probabilisticembeddingscrossmodalretrieval}. 

As a cornerstone paradigm, contrastive learning-based dual-encoder models (e.g., CLIP~\citep{clip}, BLIP~\citep{blip}, ALBEF~\citep{li2021alignfusevisionlanguage}) have achieved remarkable results by aligning heterogeneous modalities through large-scale paired data. 
% These models employ contrastive objectives to maximize similarity between matched modality pairs while minimizing it for mismatched ones. Constrained by the alignment methods between modalities, 
These models primarily focus on aligning global semantics while neglecting the fine-grained semantic correspondences between local components. This deficiency leads to suboptimal performance in downstream tasks such as visual grounding~\citep{xiao2024visualgroundingsurvey} and attribute-focused image retrieval~\citep{li2025highlightingmatterspromptableembeddings}.

Multimodal Large Language Models (MLLMs) have demonstrated significant advances in generalized vision-language understanding. Unlike CLIP-based models that encode text and images separately before alignment, MLLMs like Qwen-VL~\citep{qwen2.5-vl} and LLaVA-OneVision~\citep{li2024llavaonevisioneasyvisualtask} utilize interleaved text-image sequences as input. This approach enables direct capture of fine-grained semantic correspondences between text and images.
% achieving more precise cross-modal understanding and generation in tasks such as visual question answering, image captioning, and multimodal reasoning. How to construct embedding representations using MLLMs is one of the hot topics in the research community. Researchers are actively investigating MLLMs for unified multimodal representation learning. 
VLM2Vec~\citep{vlm2vec} proposes a contrastive training framework that transforms MLLMs into general multimodal embedding models. E5-V~\citep{jiang2024e5vuniversalembeddingsmultimodal} maps different modality inputs to a unified embedding space through specially formatted prompts, eliminating modality gaps without relying on multimodal training data. GME~\citep{zhang2025gmeimprovinguniversalmultimodal} advances multimodal embedding by introducing large-scale, high-quality synthetic multimodal datasets, significantly improving cross-modal retrieval performance in MLLMs.

While multimodal embeddings have achieved substantial performance improvements through MLLMs, these advances remain predominantly data-dependent rather than methodologically grounded. MLLMs are inherently constrained by their autoregressive next token prediction objective, which fundamentally differs from the task-related application format associated with embeddings. Contrastive learning based on massive data is not an efficient method for achieving transformation between the two task paradigms. Some studies have attempted to address this issue by proposing a pre-training stage for multimodal embeddings to achieve efficient transformation. UniME~\citep{unime} optimizes the language component embeddings in MLLMs by distilling knowledge from a text embedding model. MoCa~\citep{moca} replaces the causal attention mechanism of MLLM with bidirectional attention and designs a mask-based pre-training task to optimize the embedding learning process. The success of the above methods also heavily relies on high-quality, relevant data. In this paper, we propose a simple and effective pre-training strategy to reduce reliance on high-quality data.

We consider that a good embedding should possess two key characteristics: (1) \textbf{Comprehensive Information Coverage}: A good embedding should encompass as much of the input information as possible. (2) \textbf{Distinguishing Features}: Information relevant to matching should be highlighted within the embedding. The previous methods assumed that contrastive learning could achieve the simultaneous optimization of two objectives. Therefore, the optimization process requires a large amount of data. We attempt to decompose the optimization of them through a compressed pre-training task. During the compression pre-training phase, we divided the input into three parts: the input image, a set of learnable compression tokens, and image-based dialogue. By modifying the attention mechanism, we constrained the compressed tokens to extract information only from the image. Then we trained MLLMs to recover information from these compressed tokens, thereby completing the dialogue generation. At this stage, MLLMs are encouraged to generate comprehensive and rich compressed representations to address a wide variety of questions. During the contrastive learning phase, MLLMs focus on compressing token embeddings that are relevant to matching, thereby enhancing retrieval performance. Unlike other pre-training methods, our approach requires a smaller amount of data. The effectiveness of our proposed pre-training method hinges on whether the dialogue data is complex and diverse. To this end, we designed a high-quality data generation strategy that enables MLLM to automatically generate multi-turn dialogue data from a single image. This further reduces our reliance on data sources. Experimental results demonstrate that our approach achieves comparable performance to other pre-training methods while utilizing only approximately 10\% of the training data volume required by other pre-training methods.
Our main contributions are summarized as follows:
\begin{itemize}[leftmargin=*]
\item We propose a compressed pre-training strategy combined with downstream contrastive learning, successfully transforming MLLMs into competitive multimodal embedding models.
\item To reduce reliance on high-quality data during the pretraining phase, we propose an automated data synthesis method to supply data for our pretraining.
\item Extensive experiments show that our pre-training strategy is simple and efficient. Training MLLMs with LoRA on small datasets can achieve competitive performance. We also conducted extensive analyses to show how it takes effect.
\end{itemize}
\section{Related Work}

\subsection{Vision-Language Models for Multimodal Embedding}
Embedding models are fundamental components of numerous downstream applications, including retrieval, clustering, and classification. 
Early works like CLIP~\citep{clip}, BLIP~\citep{blip}, and their variants primarily focused on learning universal representations from large-scale image-text pairs. These models encode images and text separately, then align them in a latent space. 
As a result, they primarily focus on aligning global semantics while neglecting the fine-grained semantic correspondences between local regions. 
Multimodal large language models process data from different modalities through an interleaved text-image format. They provide powerful backbones for multimodal embedding models.
VLM2Vec~\citep{vlm2vec, vlm2vec-v2} transforms multimodal large language models (MLLMs), such as Phi-3.5-vision~\citep{phi3}, LLaVA-1.6~\citep{llava1.6}, and Qwen-VL~\citep{qwen2-vl, qwen2.5-vl}, into competitive multimodal embedding models through large-scale contrastive learning. E5-V~\citep{jiang2024e5vuniversalembeddingsmultimodal} employs specially designed prompts to project inputs from diverse modalities into a unified representation space, effectively bridging modality gaps without multimodal training data. 
GME~\citep{zhang2025gmeimprovinguniversalmultimodal} enhances multimodal embeddings by introducing large-scale, high-quality synthetic multimodal datasets, which mitigate modality imbalance in training and substantially boost the cross-modal retrieval performance of MLLMs.

\subsection{Pretraining for MultiModal Embedding}
While contrastive learning can effectively learn global alignment across modalities, it struggles to support deep cross-modal integration and fine-grained semantic understanding. To overcome this limitation, the objectives different from contrastive learning are integrated into the model's pre-training process to enhance performance.

In multimodal learning, LXMERT~\citep{lxmert} and UNITER~\citep{uniter} apply Masked Language Modeling (MLM) during pretraining to jointly learn image-text representations. Research on multimodal pre-training has also explored reconstruction-based objectives~\citep{vilt, vlmo, mme5, janus, janus-pro, seed-x}. Based on CLIP, ALBEF~\citep{li2021alignfusevisionlanguage} and ViLT~\citep{vilt} introduce Image-Text Matching and Masked Language Model tasks to learn multimodal fusion and fine-grained representations. BLIP~\citep{blip} and CoCa~\citep{yu2022cocacontrastivecaptionersimagetext} integrate the language modeling task into the pretraining process of embeddings, leveraging the model's generative capabilities to optimize embedding performance.
% Many works transfer the successful training methods in natural language understanding models, like applying Masked Language Modeling (MLM) during pretraining models such as LXMERT \citep{lxmert} and UNITER \citep{uniter} to jointly learn image-text representations, but using relatively small transformer architectures. Another line of work designs reconstruction-based objectives in multimodal settings\citep{vilt, vlmo, mme5, janus, janus-pro, seed-x}. For example, ViLT \citep{vilt} uses both MLM and Image-Text Matching (ITM) during training. 

Multimodal embedding models based on MLLMs have achieved significant progress across various evaluation tasks through large-scale contrastive learning. 
MLLMs are primarily designed to perform text generation tasks, differing from embedding applications. This gap highlights the need for designing pre-training methods tailored to embedding tasks.
UniME~\citep{unime} introduces a pretraining stage to enhance the multimodal embedding capabilities of the model. It involves pretraining with textual discriminative knowledge distillation, where knowledge is transferred from a powerful LLM-based teacher embedding model to strengthen the language component of MLLMs. 
Considering the efficiency of bidirectional attention for data encoding, MoCa~\citep{moca} introduces a modality-aware continual pre-training stage. This phase employs a joint reconstruction objective that denoises interleaved text-image inputs using both Masked Language Modeling (MLM) and Masked Autoencoding (MAE), effectively enhancing the model’s capacity for bidirectional contextual representation and cross-modal alignment.

% MoCa \citep{moca} introduces a bidirectional attention mechanism and an MLM task and designs diverse modality-aware objectives to continually pretrain a causal vision-language model into a discriminative model. However, such a stage needs huge data (approximately 30B tokens) and is resource-consuming. And it's hard to say such a stage fully utilizes the capacity of the casual pretrained model. Some works retain casual features of the pretrained model. 

% employs textual discriminative knowledge distillation from a powerful LLM-based teacher embedding model to enhance the embedding capability of the MLLM's language component. Then, an instruction-based contrastive learning stage is conducted against hard negative samples to build universal embeddings across text and vision modalities. 

% UniME \citep{unime} develops the discriminative capability of the vision-language model's language component by knowledge distillation from an LLM-based teacher embedding model. 
% However, the two modalities separately into two stages may hurt the modality fusion capability, and the textual embedding distillation stage relies on a powerful teacher embedding model and breaks the gap between the pretraining stage and the contrastive learning stage.

\section{Method}
\subsection{Preliminary}
\subsubsection{Multimodal Embedding Models} 
VLMs split images $I$ and text $T$ into patches and tokens, respectively, map them into the same embedding space through a text encoder and a visual encoder, and finally generate unified hidden state features $H\in\mathbb{R}^{L\times N\times D}$.
Most multimodal embedding models derive a holistic representation of the input from these hidden states. 
For causal models, the hidden states in each layer depend only on previous hidden states, and many works select the last token of the final layer—typically corresponding to the [EOS] token. 
In contrast, some VLM models convert causal attention into bidirectional attention, thereby gaining access to all input information. Consequently, they often use mean pooling over all sequence representations or train a special token to aggregate the final embedding.

\subsubsection{Contrastive Learning}
In the optimization of embedding models, contrastive learning serves as the core training paradigm. Its fundamental principle involves learning discriminative representations by pulling semantically similar samples closer together while pushing irrelevant samples farther apart. This paradigm has been widely applied in downstream tasks such as retrieval and recommendation systems. 
During training, each instance is structured as a tuple $(q, d^{+}, \{d^{-}_1,...,d^{-}_K\})$ where $q$ is the query, $d^+$ is a positive item,
and $\{d^{-}_1,...,d^{-}_{|K|}\}$ corresponds to a set of $K$ negative samples within the same batch. The widely used objective function in this setting is the InfoNCE~\citep{infonce}, defined as:
\begin{equation*}
\centering
\begin{aligned}
     \mathcal{L} = -\log\frac{e^{sim(h_{q}, h_{d^{+}})/\tau}}{e^{sim(h_{q}, h_{d^{+}})/\tau} + \sum_{i=1}^{K}e^{sim(h_{q}, h_{d^{-}_{i}})/\tau}},
\end{aligned}
\end{equation*}where $sim(\cdot)$ denotes the similarity function and $\tau$ is the temperature.

\definecolor{A}{RGB}{189, 255, 166}
\definecolor{C}{RGB}{244, 177, 131}

\begin{figure*}[htbp!]
    \centering
    \includegraphics[width=0.95\linewidth]{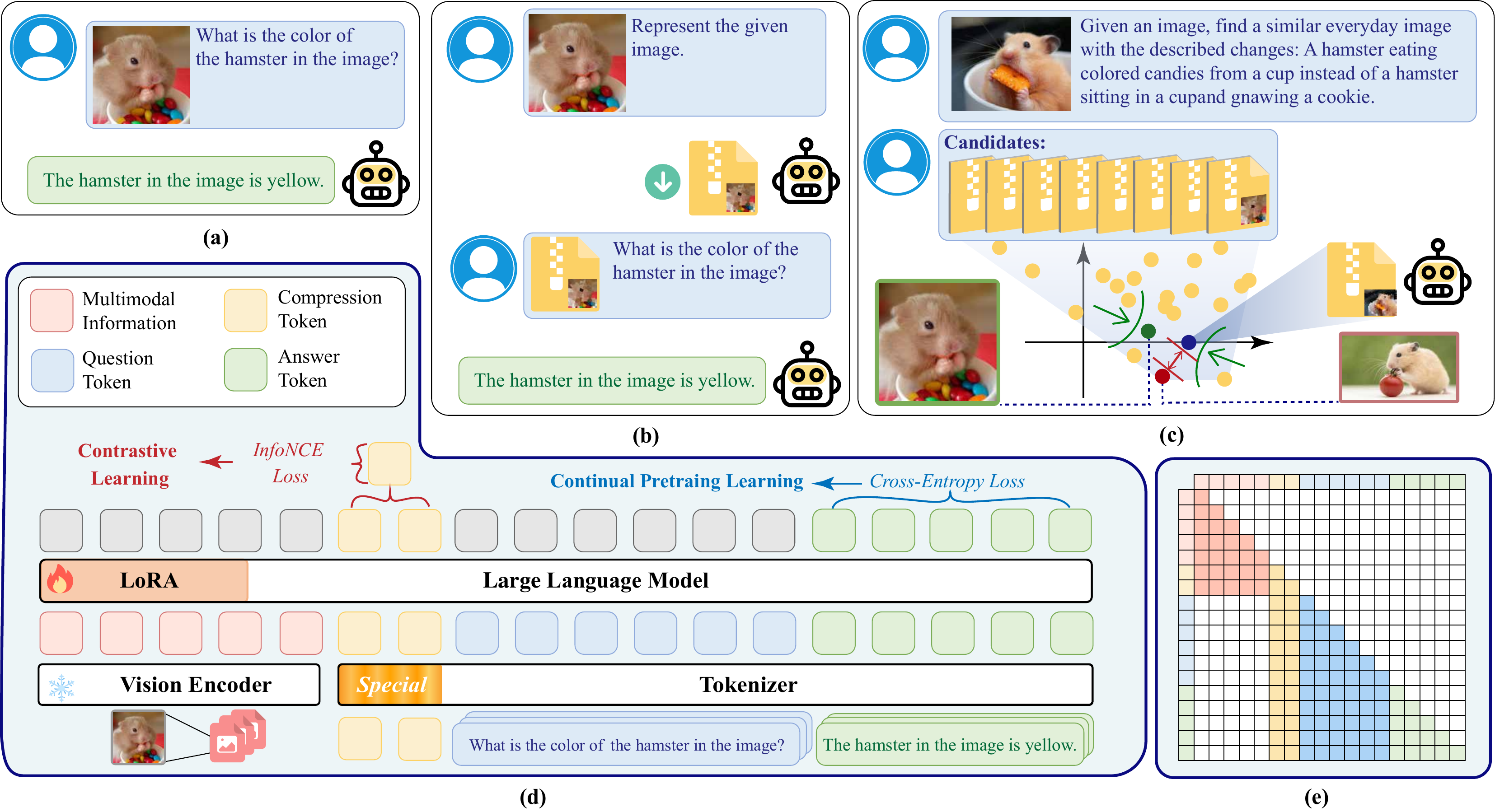}
    \caption{Architecture of \methodname, the top part demonstrates three training stages: \textbf{(a) Instruction-Tuning}, \textbf{(b) Continual PreTraining} and \textbf{(c) Contrastive Learning}. In \textbf{(a)} and \textbf{(b)}, our continual pretraining stage inherits the same format of Question-Answering (QA) task as the upstream stage, while the answer in the upstream depends on the whole image, and our stage depends on the \textbf{\textcolor{C}{Compression Tokens}} which condense the image. In \textbf{(c)}, the contrastive learning stage inherits the compression tokens as representations of multimodal input to apply to upstream retrieval-based tasks. The bottom part shows the implementation of our continual pretraining stage: \textbf{(d)} shows the model architecture where all the inputs are concatenated to calculate simultaneously, and the dependencies are driven by our modified causal attention masks in \textbf{(e)}.  }
    \label{fig:main}
\end{figure*}
% 动机
\subsection{Compression Pretraining}
We argue that a good representation should first incorporate as much of the input information as possible, then highlight the part of it related to matching. However, previous studies \citep{fastv, voco-llama} reveal substantial spatial and semantic redundancies in existing representations. To reduce these redundancies and improve representation efficiency, we designed an additional compressed pre-training stage to train multimodal embedding models to extract comprehensive information from the input.

% \subsubsection{Question-Answer based Reconstruction}
% MLLMs经过大量问答数据的训练
% 对话数据的特点，形式多变, 模型的能力取决于问题的多样性和全面性。
\subsubsection{Compression based on Question-Answer}
Question-answering data formats are widely used in training large language models due to their flexibility, particularly in the critical stages of SFT and RLHF. Benefiting from their formalization capabilities, common tasks such as classification and summarization can be uniformly converted into question-answering formats. Furthermore, some complex evaluations for MLLMs are also conducted through question-answering.

We argue that leveraging question-answering data as a supervisory signal offers a novel approach for optimizing the compression capabilities of multimodal embedding models. Given an image, when questions are diverse and complex, MLLMs must comprehensively and accurately understand the image information to provide corresponding answers. MLLMs trained through this method can serve as a superior backbone for multimodal embedding models.

\subsubsection{Compression Mechanism}
For an input image $I$ and its corresponding question $Q$ and answer $A$, we first insert a set of compression tokens $C=[C_1, ..., C_{K}]$ behind the image, where $K$ is significantly smaller than the length of the image input. The input is serialized as:
% {
% \begin{align*}
% Input =& <|image\_pad|>\ldots,<|image\_pad|>, 
% \\&<C_1>, \ldots, <C_{K}>, Question, \\&Answer\nonumber.
% \end{align*}
% }
\begin{equation*}
\centering
\scalebox{0.95}{$
\begin{aligned}
\text{Input} =& <|\text{image\_pad}|>, \ldots, <|\text{image\_pad}|>, \\
       & <C_1>, \ldots, <C_{K}>, \text{[Question]}, \text{[Answer]}. \nonumber
\end{aligned}$}
\end{equation*}

% For each multimodal input $I$, we generate some questions $Q$. We first generate corresponding answers $A$ according to the input, where the question-answer pairs are organized as a multi-turn conversation.
% \begin{align}
%     A \sim P(\ \cdot\ |\ X\oplus Q;\ \theta)
% \end{align}

% We then distill the model from itself to obtain the capability of compressing and reconstructing multimodal information. 
Unlike the training process of SFT, our pre-training process aims to maximize the following objective:
\begin{align}
    P(A\ |\ \mathcal{C}\oplus Question;\ \theta). \nonumber
\end{align}

% where $\mathcal{C}(X)$ is a sequence of hidden states which works as a compression of the multimodal input $X$. 

% To achieve compression of the multimodal input $\mathcal{C}(X)$, we first introduce a set of dedicated compression tokens $C=[C_1, ..., C_{K}]$ to encode the compressed information, where $K$ is significantly smaller than the length of the multimodal input. The compressed hidden states are computed from the input $X$ using attention mechanism. When these compression tokens $C$ are concatenated with question-answer pairs, the Attention Mask-Guided Information Compression mechanism ensures effective feature compression.

% For the compression pretraining stage, we employ the standard cross-entropy loss.

% 我们的范式通过语言模型的训练目标保留了模型的 causal 特性，通过做 QA 任务，与模型指令微调阶段对齐，最大化保留并迁移了模型在预训练阶段学习到的能力

Benefiting from casual attention, compressed tokens extract information from images to support the subsequent training of question-answering. The pretraining method we propose effectively bridges the gap between instruction models and embedding models. It not only fully leverages the knowledge stored in model parameters but also is similar in task form to SFT. Unlike SFT, which relies on high-quality question-answering data (typically requiring questions to be as complex as possible and answers to be strictly accurate) for training, our proposed compression pre-training method emphasizes comprehensive coverage and diversity of questions without strictly demanding answer accuracy.

After pretraining, the contrastive learning phase proceeds without any conversational components. We extract representations from the final hidden states corresponding to the compression tokens, apply mean pooling to aggregate these features, and utilize contrastive learning to align the multimodal embeddings.

\subsubsection{Attention Mask-Guided Information Compression}
In our implementation, the compression capability is integrated into the calculation of the QA loss. As illustrated in Figure~\ref{fig:main}(d), the multimodal input, compression tokens $C$, and conversational inputs \(Q \oplus A\) are concatenated into a single sequence. A key design consideration is the dependency structure among these three components: the compression tokens naturally depend on the multimodal input, following standard practice. Crucially, the conversational segment depends solely on the compression tokens, implying that the hidden states of the conversational part cannot be computed using information from the input segment. To enforce this dependency structure, we modify the causal attention mask. As shown in Figure \ref{fig:main}\textbf{(e)}, we mask the lower triangular region between the conversational and information segments, setting the corresponding attention scores to zero.

\subsection{Automatic Data Generation}\label{sec:data_gen}
Existing pre-training methods often exhibit strong dependence on both the quantity and diversity of training data. Our proposed pre-training paradigm leverages complex, multi-source QA data, which places higher demands on data quality.
Drawing inspiration from related work such as Self-Instruct~\citep{wang2023selfinstructaligninglanguagemodels}, we explore the potential of MLLMs to generate diverse, high-quality training samples autonomously. 

Given the retrieval-oriented characteristics of our task and the objectives of our pre-training, we avoid random instruction generation and instead prioritize the coverage and comprehensiveness of instructions with respect to image content. We employed Qwen2.5-VL-7B to generate three to five potential questions for an image at random, and instructed Qwen2.5-VL-7B to provide answers to the generated questions through a multi-turn dialogue format. Detailed instructions are provided in the Figure~\ref{sec:autoqa}, and statistical information of the pre-training data is presented in the Table~\ref{tab:data_training_stat}.
\begin{table}[htbp!]
        \centering
        \caption{Statistics of Pretraining Data.}
        \resizebox{\linewidth}{!}{
        \begin{tabular}{l|rrr|r}
            \hline
            \diagbox{Dataset}{\# Turns} & 3 & 4 & 5 & Total  \\
            \hline
            CIRR & 79 & 138 & 16,423 & 16,640 \\
            HatefulMemes & 74 & 26 & 8,400 & 8,500 \\
            MSCOCO & 706 & 399 & 2,2507 & 2,3612 \\
            MSCOCO\_i2t & 105 & 64 & 29,830 & 29,999\\
            MSCOCO\_t2i & 115 & 54 & 29,831 & 30,000 \\
            N24News & 59 & 42 & 29,899 & 30,000 \\
            SUN397 & 9 & 14 & 19,827 & 19,850 \\
            VOC2007 & 227 & 156 & 6,293 & 6,676 \\
            Visual7W & 23 & 25 & 14,318 & 14,366 \\
            WebQA & 17 & 7 & 12,849 & 12,873 \\  
            \hline
            Total & 1,739 & 1,295 & 219,482 & 222,516 \\
            \hline
        \end{tabular}}
        \label{tab:data_training_stat}
\end{table}

\section{Experiments}
\subsection{Datasets}
Both the pre-training and contrastive learning data for \methodname~originate from the MMEB-V1~\citep{vlm2vec}, which comprises 36 datasets categorized into 4 meta-tasks: classification, visual question answering, retrieval, and visual grounding. 
We randomly sampled approximately 220K examples from the MMEB-V1 training set and constructed a pre-training dataset based on the images it contained, following the method described in Section~\ref{sec:data_gen}. Statistical information about the pre-training dataset is shown in Appendix Table~\ref{tab:data_training_stat}. After pretraining is completed, we use only the training set of MMEB-V1 for contrastive learning.

\begin{table*}[t]
\centering
\caption{\textbf{Results on MMEB-V1.} ``IND'' denotes in-distribution, and ``OOD'' refers to out-of-distribution. \textbf{Bold} and \underline{underline} indicate the optimal and suboptimal performance, respectively.}
\resizebox{\textwidth}{!}{
\begin{tabular}{lrrrrrrrr} 
\hline
\multirow{2}{*}{\textbf{Models }} & \multirow{2}{*}{\textbf{\#Params}} & \multicolumn{4}{c}{\textbf{Per Meta-Task Score }} & \multicolumn{3}{c}{\textbf{Average Score}} \\ 
\cline{3-9}
 &  & Classification & VQA & Retrieval & Grounding & IND & OOD & Overall \\ 
\hline
\# of Datasets $\to$ &  & 10 & 10 & 12 & 4 & 20 & 16 & 36 \\ 
\hline
CLIP (ViT-L) & 0.4B & 55.2 & 19.7 & 53.2 & 62.2 & 47.6 & 42.8 & 45.4 \\
OpenCLIP (ViT-L) & 0.4B & 41.5 & 6.9 & 44.6 & 53.5 & 32.8 & 36.0 & 36.6 \\
GME (Qwen2-VL) & 2B & 56.9 & 41.2 & 67.8 & 53.4 &- &- &- \\
UNITE (Qwen2-VL) & 2B & 63.2 & 55.9 & 65.4 & 75.6 & 65.8 & 60.1 & 63.3 \\
VLM2Vec (Qwen2.5-VL) & 3B & 55.3 & 57.3 & 62.7 & 73.2 & - & - & 60.3 \\
E5-V (LLaVA-1.6) & 7B & 39.7 & 10.8 & 39.4 & 60.2 & 34.2 & 33.9 & 33.9 \\
MMRet (LLaVA-1.6) & 7B & 56.0 & 57.4 & 69.9 & 83.6 & 68.0 & 59.1 & 65.8 \\
VLM2Vec (Qwen2-VL) & 7B & 62.6 & 57.8 & 69.9 & 81.7 & 72.2 & 57.8 & 65.8 \\
CAFe (LLaVA-OV) & 7B & 65.2 & \underline{65.6} & 70.0 & \underline{91.2} & \textbf{75.8} & 62.4 & 69.8 \\
UNITE (Qwen2-VL) & 7B & \textbf{68.3} & 65.1 & 71.6 & 84.8 & 73.6 & 66.3 & 70.3 \\
mmE5 (Llama-3.2-Vision) & 11B & \underline{67.6} & 62.8 & 70.9 & 89.7 & 72.3 & \underline{66.7} & 69.8 \\ 
\hline
UniME (Phi3.5-V) & 4.2B & 54.8 & 55.9 & 64.5 & 81.8 & 68.2 & 52.7 & 64.2 \\
UniME (LLaVA-1.6) & 7B & 60.6 & 52.9 & 67.9 & 85.1 & 68.4 & 57.9 & 66.6 \\
MoCa (Qwen2.5-VL) & 3B & 59.8 & 62.9 & 70.6 & 88.6 & 72.3 & 61.5 & 67.5 \\
MoCa (Qwen2.5-VL) & 7B & 65.8 & 64.7 & \textbf{75.0} & \textbf{92.4} & 74.7 & \textbf{67.6} & \underline{71.5} \\
\rowcolor{gray!20}
\rowcolor{gray!20} \methodname~(Qwen2.5-VL) & 3B & 61.3 & 65.1 & 70.0 & 82.7 & 71.3 & 61.6 & 67.5 \\
\rowcolor{gray!20} \methodname~(Qwen2.5-VL) & 7B & 67.4 & \textbf{70.6} & \underline{72.4} & 87.6 & \underline{75.2} & \textbf{67.6} & \textbf{72.2} \\
\hline
\end{tabular}
}
\label{tab:results}
\end{table*}

\subsection{Training Procedure}
\label{sec:exp_set}
We employ the Qwen2.5-VL as the backbone for our multimodal embedding model.
Our training procedure consists of two stages: compression pre-training followed by contrastive learning. To handle inputs from multiple modalities and accommodate images of varying sizes, we employ dynamic resolution via MRoPE \citep{qwen2-vl}, limiting the maximum number of vision tokens to 1024. The number of compression tokens is 32. Due to computational resource constraints, we set the batch size to 256 during the pretraining phase. In the contrastive learning phase, we scaled it up to 1024 using the GradCache~\citep{gao2021scalingdeepcontrastivelearning}. For all experiments, we used LoRA~\citep{lora} (rank=16) and a learning rate of 5e-5 in both stages, and our GPU requirements are only one-quarter of those for MoCa~\citep{moca}.
%all experiments are conducted on 8$\times$NVIDIA A800 (80GB) GPUs.

\subsection{Evaluation and Metrics}
We evaluate the performance of \methodname~on the MMEB-V1 benchmark, which provides evaluation benchmarks across four meta-tasks consistent with those in the training set. The benchmark comprises 36 evaluation datasets, including 20 in-distribution and 16 out-of-distribution subsets. We employ Precision@1 as our evaluation metric, emphasizing top-ranked results to reflect their applicability in real-world scenarios.

\section{Overall Performance}
% 预训练/不预训练 说明预训练的有效性
We compared the performance of \methodname~against other competitive multimodal embedding models on the MMEB. The results are shown in Table~\ref{tab:results}. We categorize these baselines into two groups: one directly employs contrastive learning, while the other incorporates an additional pre-training stage.
% 性能表现
For the same backbone, introducing additional pre-training can effectively improve its retrieval performance. This highlights the importance of the pre-training stage when converting MLLMs into multimodal embedding models. 
Compared to other pre-training approaches, the compressed pre-training stage proposed in \methodname~demonstrates superior effectiveness. Experimental results show that \methodname~achieves optimal or near-optimal levels across multiple key metrics, further validating the effectiveness of the compressed pre-training strategy.
% 训练效率
In terms of training efficiency, \methodname~employs the LoRA for training. Compared to the best baseline MoCa, \methodname~utilizes only 300 million tokens during the pre-training phase, significantly fewer than the 30 billion tokens required by MoCa. Furthermore, during the contrastive learning phase, \methodname~achieves state-of-the-art performance using only half the training data of MoCa and with a smaller batch size. This demonstrates that \methodname~is both simple and efficient, significantly reducing computational resource requirements without sacrificing performance. 

\section{Further Analysis}
\subsection{Scaling of Compression Tokens}\label{abla_token}
\begin{figure*}[htbp!]
    \centering
    \includegraphics[width=\linewidth]{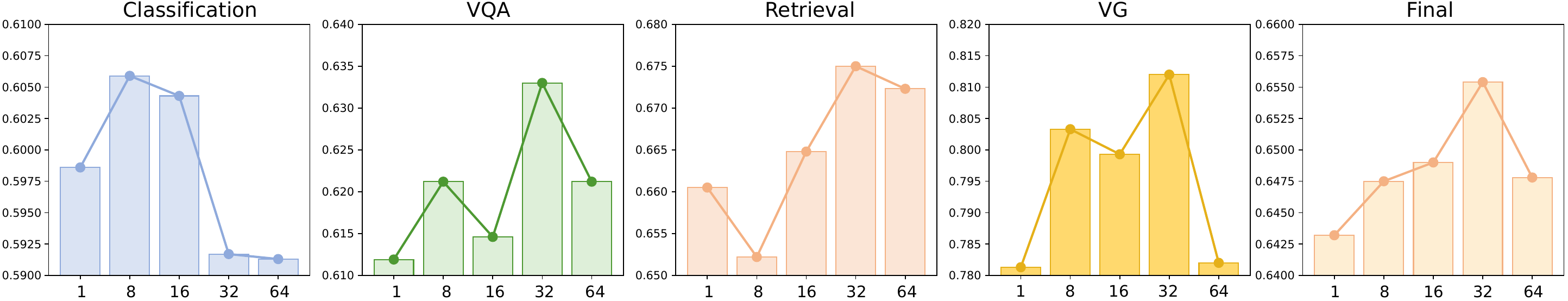}
    \caption{The impact of different numbers of compressed tokens on performance.}
    \label{fig:c_num}
\end{figure*}
We utilised 32 compressed tokens to extract information from the multimodal input. To investigate the impact of the number of tokens on performance, we adjusted the number of compressed tokens in \methodname, retrained the multimodal embedding model, and compared performance across different tasks. We employed Qwen2.5-VL-3B as the backbone of \methodname~for analysis. 
Considering training efficiency, we only used 500K training samples randomly sampled from the MMEB-V1 during the contrastive learning phase. The results are shown in Figure~\ref{fig:c_num}. 

\begin{figure}[htbp]
    \centering
    \begin{subfigure}{0.29\linewidth}
        \centering
        \includegraphics[width=\linewidth]{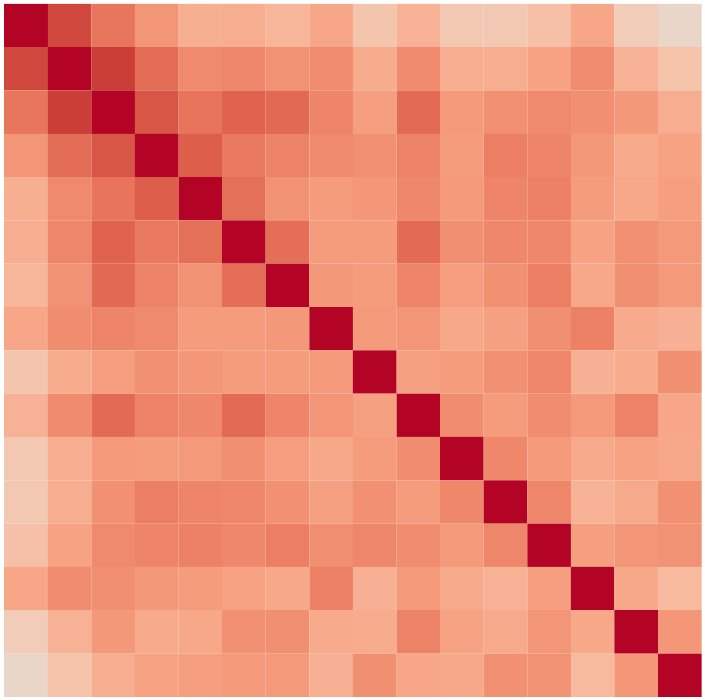}
        \caption{}
        \label{fig:sim16}
    \end{subfigure}
    \begin{subfigure}{0.29\linewidth}
        \centering
        \includegraphics[width=\linewidth]{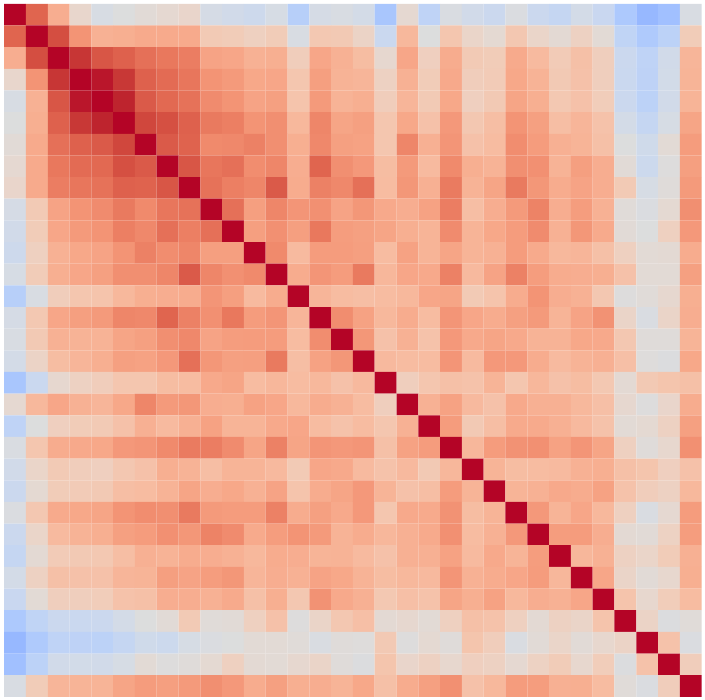}
        \caption{}
        \label{fig:sim32}
    \end{subfigure}
    \begin{subfigure}{0.36\linewidth}
        \centering
        \includegraphics[width=\linewidth]{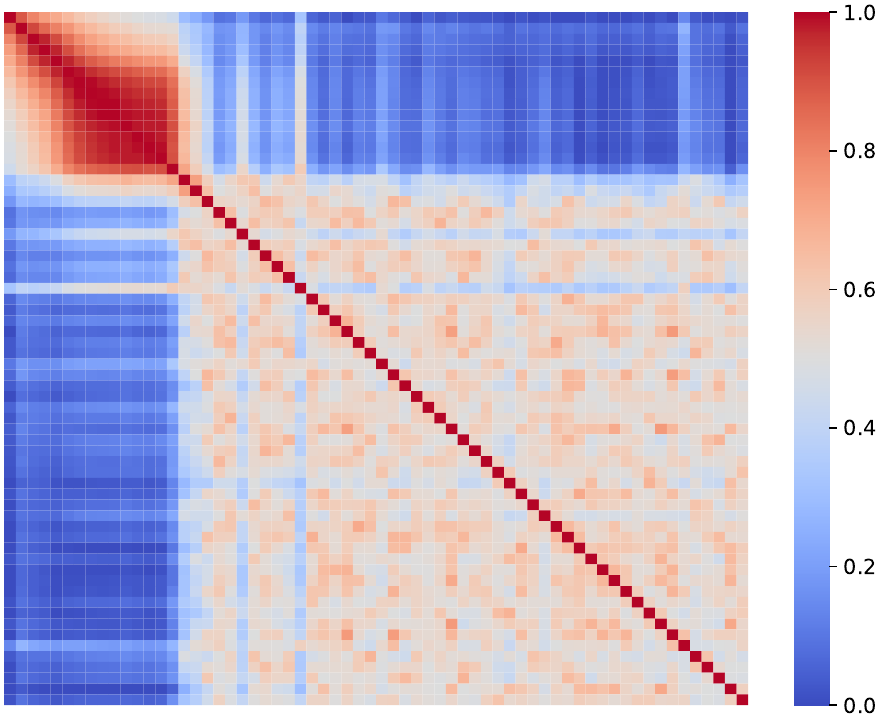}
        \caption{}
        \label{fig:sim64}
    \end{subfigure}
    \caption{Similarity between compression tokens with different numbers: (a) 16, (b) 32, and (c) 64.}
    \label{fig:sim}
\end{figure}

\definecolor{lightblue}{RGB}{160, 216, 255}
\definecolor{darkblue}{RGB}{0, 0, 255}
We found that, regardless of the number of compressed tokens used, \methodname~outperforms the baseline (VLM2Vec based on the same backbone) without additional pretraining. This demonstrates the effectiveness of the compressed pre-training strategy. As the number of compressed tokens increases, CoMa's average performance across different tasks follows an initial upward trend followed by a decline: performance gradually improves in the initial phase, but beyond 32 tokens, performance actually decreases as the token count increases. 
It is evident that as the number of compression tokens increases, the capacity of the compression space expands, thereby enabling the compression tokens to incorporate a greater amount of information. To investigate why performance declines as the number of compressed tokens decreases, we evaluated the pairwise similarity between compressed tokens and presented the results in Figure~\ref{fig:sim}. They are calculated on 100 randomly sampled instances. 
Compared to 16 tokens, 32 tokens provide sufficient space for compressing input information. The compression space offered by 64 tokens contains redundant information (\textcolor{darkblue}{Dark Blue Area}) that may interfere with matching, leading to performance degradation. The number of compressed tokens is closely related to the amount of data.
% \subsection{Scaling of Pre-training Data}

\subsection{The Impact of Different Pretraining Methods}

% 我们希望对于输入的信息，模型能尽可能压缩充分的信息，从而获取到信息通用的表示以完成不同的下游任务。对于 qa 任务，我们希望找到一种形式，能更加充分利用信息。如，对于同样的图片信息，我们希望问题尽可能覆盖图中充分的信息。
% TODO: general or universal
% We expect \methodname~to compress multimodal information as thoroughly as possible to obtain a general representation to cope with various and unknown downstream tasks. 
% We hope to find a format that drives the model to capture the information fully.
% There are multiple ways to achieve this function, depending on how the data is organized and the training method employed. 

% 为此，我们就图片与问题的数量对应关系，设置了几组消融实验，包括：一对多的多轮对话和拆分为多个单轮的形式、一对一的单轮对话（随机从多轮对话中抽出一个问题）
To explore what data formats and training methods enable \methodname~to learn compression capabilities more effectively, we designed several ablation experiments concerning the quantity mapping between images and questions, including: a multi-turn conversation format (one image to multiple questions) and its variant, the single-turn conversation format (one image to one question). In addition, we altered the content of the question-answer pairs, replacing them with descriptions and captions to analyse the impact of different content types on performance. In terms of training methods, we employed KL as an alternative to the cross-entropy loss in Next Token Prediction for comparison.
Regarding the experimental setting, for contrastive learning, we only used 500K training samples consistent with the ablation experiments in Section~\ref{abla_token}, and other parameter settings were kept the same as the main experiment in Section~\ref{sec:exp_set}. The results under different settings are shown in Table~\ref{tab:ablation}.

% For Question-Answering (QA) tasks,  Given a single image, for example, we expect to generate questions that cover as much information about the image as possible.
\begin{table}[htbp!]
    \centering
    \caption{Performance under different formats and training methods. Optimal results are displayed in \textbf{bold}, and the suboptimal results are shown with \underline{underlining}.}
    \resizebox{\linewidth}{!}{
    \begin{tabular}{lcccccc}
    \hline
    \multirow{2}{*}{\textbf{Format}} & \multirow{2}{*}{\textbf{Loss}} & \multicolumn{4}{c}{\textbf{Per Meta-Task}} & \textbf{Avg} \\ 
    \cline{3-6}
    & & Classification & VQA & Retrieval & VG &  \\
    \hline
    Multi Turn          & \multirow{5}{*}{CE} & 59.2 & \textbf{63.3} & 67.5 & \textbf{81.2} & \textbf{65.5} \\
    Single Turn         &                               & 59.6 & 62.2 & 66.3 & 79.6 & 64.8 \\
    Description   &                               & 59.9 & 61.6 & 66.7 & 80.8 & 64.9 \\
    Caption             &                               & \textbf{60.7} & 62.1 & \textbf{67.5} & 77.2 & \underline{65.2} \\
    \hline
    Multi Turn          & KL                           & 58.1 & 61.7 & 67.4 & 77.6 & 64.4 \\
    \hline
    \end{tabular}}
    \label{tab:ablation}
\end{table}
We found that using a multi-turn dialogue format yields better results than a single-turn dialogue format. A reasonable hypothesis is that compression differs from instruction tuning in that compression is inherently a lossy process, which means some detailed information must be discarded during compression. Under this premise, setting single questions may lead to excessive focus on the detailed aspects of the image. While multiple questions focus on different aspects of the same image, the model may autonomously balance what needs to be compressed and what can be discarded to minimise the information loss. We also found that replacing dialogue with detailed image descriptions or captions did not yield satisfactory results. This comparison demonstrates the critical importance of information coverage for pre-training. For massive datasets, image caption serves as a suboptimal yet efficient alternative solution.

\subsection{Effect of Compression Pretraining}
\begin{figure}[hbtp!]
    \begin{subfigure}{0.32\linewidth}
        \centering
        \includegraphics[width=\linewidth]{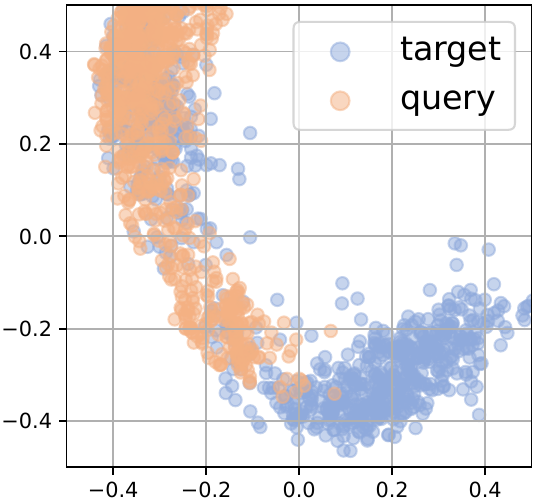}
        \caption{}
        \label{fig:train0}
    \end{subfigure}
    \begin{subfigure}{0.32\linewidth}
        \centering
        \includegraphics[width=\linewidth]{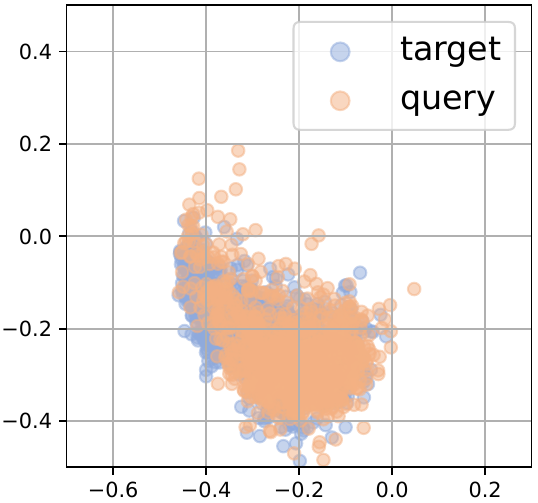}
        \caption{}
        \label{fig:train1}
    \end{subfigure}
    \begin{subfigure}{0.32\linewidth}
        \centering
        \includegraphics[width=\linewidth]{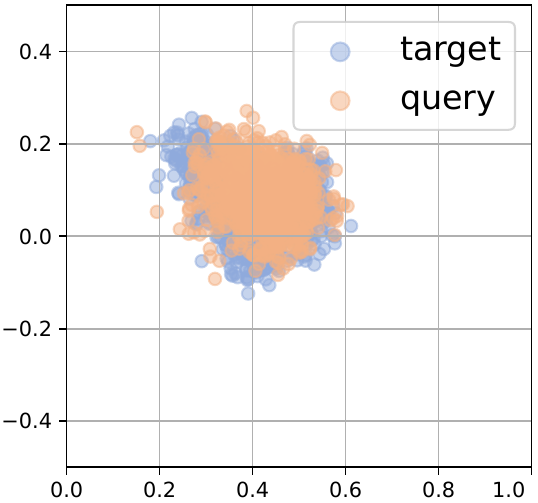}
        \caption{}
        \label{fig:train2}
    \end{subfigure}
    \caption{Representations of queries and targets across Three Stages: (a) Base, (b) Compression Pretraining, and (c) Contrastive Learning. Representations in (a) are extracted from the base model via the [EOS] token, while the others are via compression tokens. The representations are decomposed using Principal Component Analysis (PCA).}
    \label{fig:reprs}
\end{figure}

The compression pretraining strategy we propose is close to supervised fine-tuning in the training paradigm, while it is similar to embedding models in terms of functionality. 
To explore the underlying mechanisms of compression pre-training, we analyzed how representations evolve for the same input across different training stages and visualized these changes in Figure~\ref{fig:reprs}. This result provides an intuitive illustration of the model's representational evolution.
% 为什么压缩是好的中间过程
% Our compression pretraining stage inherits the task format from upstream stages. For the downstream contrastive learning stage, our compression representations learned via language modeling objectives provide a starting point for the retrieval-oriented representations, which makes it easier to fit in the contrastive learning. As shown in figure \ref{fig:reprs}, after our continual pretrained stage, the representations are much more compact than the initial base model and already have the shape and size that is close to the contrastive learning trained one. In the contrastive learning stage, the model learns to reshape the distribution constrained in a smaller space and just shift the intermediate representations to the target space, which is easier to fit.

It is evident that after compressed pre-training, the representation of the same input becomes closer to the representation obtained after final contrastive learning. The compressed pre-training method bridges the gap between instruction models and embedding models, effectively reducing the fitting difficulty in contrastive learning.

\subsection{Is Distillation Better For Compression?}
% \section{Case Study}
% 与普通的指令微调不同，我们基于蒸馏的形式——更确切的，自蒸馏——似乎基于蒸馏的方式，如使用 KL 散度 loss 会获得更好的性能。故我们设计了基于 KL 散度的损失，如下

Unlike standard instruction-tuning, our approach focuses on optimising the compression tokens in the latent space. It seems to work better when using distillation training methods, such as applying the KL divergence loss, as it allows for fine-grained optimisation. However, as shown in the Table~\ref{tab:ablation}, the training based on cross-entropy loss actually brings better performance. Intuitively, KL divergence requires complete consistency between output distributions, which seems overly strict supervision for inherently lossy tasks like compression. What we need is to strike a balance between compression capability and generalisation ability, but the KL criterion may limit the generalisation capability of the model. We cannot require the distribution of the model's output to be consistent with that before compression; instead, we can only supervise the expected answers, which is achieved by the standard cross-entropy loss.

\begin{figure}[htbp]
    \centering
    \includegraphics[width=\linewidth]{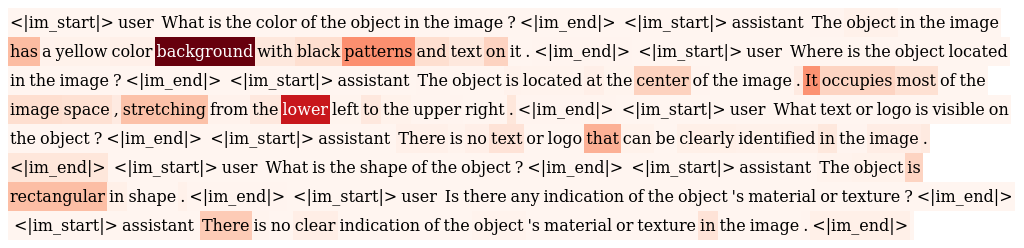}
    \includegraphics[width=\linewidth]{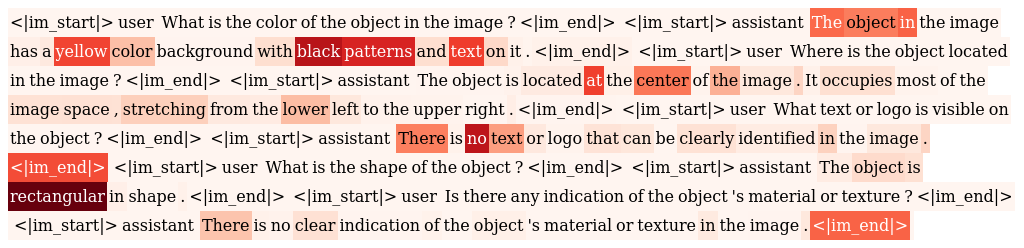}
    \caption{Loss distribution across tokens for Cross-Entropy (Top) and KL Divergence (Bottom). }
    \label{fig:loss_dist}
\end{figure}

The Figure~\ref{fig:loss_dist} shows an example of training loss distribution under the same settings and instance. It can be seen that both cross-entropy and KL divergence focus most of their loss on the similar tokens (those related to the answer). In comparison, cross-entropy is more concentrated, while KL divergence distributes considerable loss to more tokens, where many of them are unimportant for compression, such as [EOS].

% TODO: 模型知道自己在压缩什么吗？是压缩字面上的信息，还是经过推理后的信息？

\section{Conclusion}
In this paper, we propose \methodname, which decouples the compression and matching functionalities within multimodal embedding models by introducing an additional compression pre-training process. Experiments demonstrate that \methodname~is both straightforward and effective, achieving state-of-the-art performance on MMEB-V1. The pre-training process of CoMa utilises only images as compressed input. However, CoMa is not limited to processing images. It can also handle multimodal data such as plain text and video. We will explore the impact of compressing different multimodal data on \methodname's performance in future work.
\section*{Limitations}
Constrained by training resources, \methodname~was pre-trained and contrastively learned only on a limited amount of data. Its performance upper bound remains to be further explored following subsequent increases in data scale.
\section*{Acknowledgments}
This work was funded by New Generation Artificial Intelligence-National Science and Technology Major Project of No. 2025ZD0123301, the National Natural Science Foundation of China (NSFC) under Grants No. 62302486 and No. 62441229, the Innovation Project of ICT CAS under Grants No. E361140, and the Strategic Priority Research Program of the CAS under Grants No. XDB0680102.

% This document has been adapted
% by Steven Bethard, Ryan Cotterell and Rui Yan
% from the instructions for earlier ACL and NAACL proceedings, including those for
% ACL 2019 by Douwe Kiela and Ivan Vuli\'{c},
% NAACL 2019 by Stephanie Lukin and Alla Roskovskaya,
% ACL 2018 by Shay Cohen, Kevin Gimpel, and Wei Lu,
% NAACL 2018 by Margaret Mitchell and Stephanie Lukin,
% Bib\TeX{} suggestions for (NA)ACL 2017/2018 from Jason Eisner,
% ACL 2017 by Dan Gildea and Min-Yen Kan,
% NAACL 2017 by Margaret Mitchell,
% ACL 2012 by Maggie Li and Michael White,
% ACL 2010 by Jing-Shin Chang and Philipp Koehn,
% ACL 2008 by Johanna D. Moore, Simone Teufel, James Allan, and Sadaoki Furui,
% ACL 2005 by Hwee Tou Ng and Kemal Oflazer,
% ACL 2002 by Eugene Charniak and Dekang Lin,
% and earlier ACL and EACL formats written by several people, including
% John Chen, Henry S. Thompson and Donald Walker.
% Additional elements were taken from the formatting instructions of the \emph{International Joint Conference on Artificial Intelligence} and the \emph{Conference on Computer Vision and Pattern Recognition}.

\bibliography{custom}
\appendix
\newpage
% \section{Statistics of Pretraining Data}
% \label{sec:appendix}
% \begin{table}[htbp!]
%         \centering
%         \caption{Statistics of Pretraining Data.}
%         \resizebox{0.9\linewidth}{!}{
%         \begin{tabular}{l|rrr|r}
%             \hline
%             \diagbox{Dataset}{\# Turns} & 3 & 4 & 5 & Total  \\
%             \hline
%             CIRR & 79 & 138 & 16,423 & 16,640 \\
%             HatefulMemes & 74 & 26 & 8,400 & 8,500 \\
%             MSCOCO & 706 & 399 & 2,2507 & 2,3612 \\
%             MSCOCO\_i2t & 105 & 64 & 29,830 & 29,999\\
%             MSCOCO\_t2i & 115 & 54 & 29,831 & 30,000 \\
%             N24News & 59 & 42 & 29,899 & 30,000 \\
%             SUN397 & 9 & 14 & 19,827 & 19,850 \\
%             VOC2007 & 227 & 156 & 6,293 & 6,676 \\
%             Visual7W & 23 & 25 & 14,318 & 14,366 \\
%             WebQA & 17 & 7 & 12,849 & 12,873 \\  
%             \hline
%             Total & 1,739 & 1,295 & 219,482 & 222,516 \\
%             \hline
%         \end{tabular}}
%         \label{tab:data_training_stat}
% \end{table}

\newpage
\begin{table*}[tbp!]
\centering
\caption{\textbf{Detailed MMEB-V1 Results.} Performance of baselines and \methodname~variants across 20 in-distribution (IND) and 16 out-of-distribution (OOD) datasets. OOD datasets are highlighted with a yellow background.}
\label{tab:app_mmeb_per_task}
\resizebox{\textwidth}{!}{
\begin{tabular}{lcccccccc} 
\toprule
\rowcolor[rgb]{0.851,0.851,0.851}  & \textbf{CLIP} & \textbf{VLM2Vec} & \textbf{MMRet} & \textbf{UniME} & \textbf{mmE5} & \textbf{MoCa-7B} & \textbf{\methodname~-3B} & \textbf{\methodname~-7B} \\ 
\midrule
\rowcolor[rgb]{1,0.851,0.702} \textbf{Classification (10 tasks)} &  &  &  &  &  &  &  &  \\
ImageNet-1K & 55.8 & 74.5 & 58.8 & 71.3 & 77.8 & 78.0 & 79.2 & 82.0 \\
N24News & 34.7 & 80.3 & 71.3 & 79.5 & 81.7 & 81.5 & 79.0 & 79.6 \\
HatefulMemes & 51.1 & 67.9 & 53.7 & 64.6 & 64.2 & 77.6 & 68.5 & 70.9 \\
VOC2007 & 50.7 & 91.5 & 85.0 & 90.4 & 91.0 & 90.0 & 80.4 & 83.1 \\
SUN397 & 43.4 & 75.8 & 70.0 & 75.9 & 77.7 & 76.8 & 70.8 & 76.7 \\
\rowcolor[rgb]{1,0.992,0.851} Place365 & 28.5 & 44.0 & 43.0 & 45.6 & 43 & 43.0 & 37.8 & 46.3 \\
\rowcolor[rgb]{1,0.992,0.851} ImageNet-A & 25.5 & 43.6 & 36.1 & 45.5 & 56.3 & 52.7 & 43.8 & 54.8 \\
\rowcolor[rgb]{1,0.992,0.851} ImageNet-R & 75.6 & 79.8 & 71.6 & 78.4 & 86.3 & 83.0 & 81.0 & 85.2 \\
\rowcolor[rgb]{1,0.992,0.851} ObjectNet & 43.4 & 39.6 & 55.8 & 36.4 & 62.5 & 45.2 & 56.2 & 67.3 \\
\rowcolor[rgb]{1,0.992,0.851} Country-211 & 19.2 & 14.7 & 14.7 & 18.7 & 35.4 & 30.4 & 21.2 & 28.4 \\
\textit{All Classification} & 42.8 & 61.2 & 56.0 & 60.6 & 67.6 & 65.8 & 61.8 & 67.4 \\ 
\midrule
\rowcolor[rgb]{0.702,0.702,1} \textbf{VQA (10 tasks)} &  &  &  &  &  &  &  &  \\
OK-VQA & 7.5 & 69.0 & 73.3 & 68.3 & 67.6 & 36.9 & 64.7 & 71.4 \\
A-OKVQA & 3.8 & 54.4 & 56.7 & 58.7 & 56.1 & 57.1 & 54.6 & 62.5 \\
DocVQA & 4.0 & 52.0 & 78.5 & 67.6 & 90.3 & 94.3 & 94.0 & 95.9 \\
InfographicsVQA & 4.6 & 30.7 & 39.3 & 37.0 & 56.5 & 77.2 & 74.2 & 80.2 \\
ChartQA & 1.4 & 34.8 & 41.7 & 33.4 & 50.5 & 69.8 & 67.3 & 75.0 \\
Visual7W & 4.0 & 49.8 & 49.5 & 51.7 & 51.9 & 58.5 & 54.1 & 58.1 \\
\rowcolor[rgb]{1,0.992,0.851} ScienceQA & 9.4 & 42.1 & 45.2 & 40.5 & 55.8 & 59.2 & 46.6 & 57.5 \\
\rowcolor[rgb]{1,0.992,0.851} VizWiz & 8.2 & 43.0 & 51.7 & 42.7 & 52.8 & 46.2 & 51.4 & 54.7 \\
\rowcolor[rgb]{1,0.992,0.851} GQA & 41.3 & 61.2 & 59.0 & 63.6 & 61.7 & 71.6 & 56.0 & 65.2 \\
\rowcolor[rgb]{1,0.992,0.851} TextVQA & 7.0 & 62.0 & 79.0 & 65.2 & 83.3 & 75.8 & 78.7 & 85.4\\
\textit{All VQA} & 9.1 & 49.9 & 57.4 & 52.9 & 62.6 & 64.7 & 64.2 & 70.6 \\ 
\midrule
\rowcolor[rgb]{0.702,1,0.702} \textbf{Retrieval (12 tasks)} &  &  &  &  &  &  &  &  \\
VisDial & 30.7 & 80.9 & 83.0 & 79.7 & 74.1 & 84.5 & 81.0 & 82.0 \\
CIRR & 12.6 & 49.9 & 61.4 & 52.2 & 54.7 & 53.4 & 58.7 & 60.8 \\
VisualNews\_t2i & 78.9 & 75.4 & 74.2 & 74.8 & 77.6 & 78.2 & 74.5 & 77.8 \\
VisualNews\_i2t & 79.6 & 80.0 & 78.1 & 78.8 & 83.3 & 83.1 & 77.5 & 79.3 \\
MSCOCO\_t2i & 59.5 & 75.7 & 78.6 & 74.9 & 76.4 & 79.8 & 73.6 & 77.0 \\
MSCOCO\_i2t & 57.7 & 73.1 & 72.4 & 73.8 & 73.2 & 73.9 & 72.7 & 75.1 \\
NIGHTS & 60.4 & 65.5 & 68.3 & 66.2 & 68.3 & 66.7 & 65.6 & 67.6 \\
WebQA & 67.5 & 87.6 & 90.2 & 89.8 & 88.0 & 91.4 & 88.8 & 90.3 \\
\rowcolor[rgb]{1,0.992,0.851} FashionIQ & 11.4 & 16.2 & 54.9 & 16.5 & 28.8 & 28.9 & 21.5 & 26.4 \\
\rowcolor[rgb]{1,0.992,0.851} Wiki-SS-NQ & 55.0 & 60.2 & 24.9 & 66.6 & 65.8 & 82.7 & 66.4 & 64.1 \\
\rowcolor[rgb]{1,0.992,0.851} OVEN & 41.1 & 56.5 & 87.5 & 55.7 & 77.5 & 80.4 & 70.0 & 77.3 \\
\rowcolor[rgb]{1,0.992,0.851} EDIS & 81.0 & 87.8 & 65.6 & 86.2 & 83.7 & 96.9 & 86.0 & 91.0 \\
\textit{All Retrieval} & 53.0 & 67.4 & 69.9 & 67.9 & 71.0 & 75.0 & 69.7 & 72.4 \\ 
\midrule
\rowcolor[rgb]{0.925,0.702,0.776} \textbf{Visual Grounding (4 tasks)} &  &  &  &  &  &  &  &  \\
MSCOCO & 33.8 & 80.6 & 76.8 & 76.5 & 53.7 & 84.6 & 69.4 & 73.2 \\
\rowcolor[rgb]{1,0.992,0.851} RefCOCO & 56.9 & 88.7 & 89.8 & 89.3 & 92.7 & 94.0 & 90.0 & 94.8 \\
\rowcolor[rgb]{1,0.992,0.851} RefCOCO-matching & 61.3 & 84.0 & 90.6 & 90.6 & 88.8 & 95.5 & 92.6 & 93.5 \\
\rowcolor[rgb]{1,0.992,0.851} Visual7W-pointing & 55.1 & 90.9 & 77.0 & 84.1 & 92.3 & 95.3 & 85.0 & 88.9 \\
\textit{All Visual Grounding} & 51.8 & 86.1 & 83.6 & 85.1 & 89.6 & 92.4 & 84.3 & 87.6 \\ 
\midrule
\rowcolor[rgb]{0.851,0.953,0.992} \textbf{Final Score (36 tasks)} &  &  &  &  &  &  &  &  \\
All & 37.8 & 62.9 & 64.1 & 66.6 & 69.8 & 71.5 & 67.5 & 72.2 \\
All IND & 37.1 & 67.5 & 59.1 & 68.4 & 72.3 & 74.7 & 71.6 & 75.2 \\
All OOD & 38.7 & 57.1 & 68.0 & 57.9 & 66.7 & 67.6 & 61.5 & 67.6 \\
\bottomrule
\end{tabular}
}
\end{table*}

% \newpage
% \section{Prompt for Data Generation} \label{sec:autoqa}
\begin{figure}[htbp!]  
\begin{tcolorbox}[colback=white, colframe=black, sharp corners=southwest, boxrule=0.5pt]
\textbf{Instruction} \\
You are an advanced AI assistant trained to analyze images and generate meaningful questions that capture their most important information. Analyze the given image and generate 3-5 specific questions that capture its most important visual information for retrieval purposes. \\
Each question should: \\
1. Focus on distinct key elements (objects, actions, settings). \\
2. Be clear and answerable from visual content alone. \\
3. Avoid subjective interpretations. \\
Consider, but not limited to the following questions: \\
1. Main objects and their attributes (type, color, position) \\
2. The Scene context (time/weather if apparent, location) \\
3. Visible text/logos \\
4. Notable relationships between elements \\
\textbf{[Output]}: 
\end{tcolorbox}
\caption{Prompt for Data Generation \label{sec:autoqa}}
\end{figure}

% \section{Example Appendix}

% This is an appendix.

\end{document}